# Proposal of Pattern Recognition as a necessary and sufficient principle to Cognitive Science

Gilberto de Paiva
Sao Paulo – Brazil (May 2011)
gilbertodpaiva@gmail.com

**Abstract.** *Despite the prevalence of the Computational Theory of Mind and the Connectionist Model, the establishing of the key principles of the Cognitive Science are still controversy and inconclusive. This paper proposes the concept of PATTERN RECOGNITION as NECESSARY AND SUFFICIENT PRINCIPLE for a general cognitive science modeling, in a very ambitious scientific proposal. A formal physical definition of the pattern recognition concept is also proposed to solve many key conceptual gaps on the field.*

## 1. Introduction

Are there any scientific principle, system, property, or technique that can explain the overall cognitive phenomena?

The Computational Theory of Mind and the Connectionist Model are the most successful cognitive science modeling principles, but there is still many unsolved questions about the nature of mind. This article proposes the concept of Pattern Recognition to be a formal answer to most of those questions, and build a solid Cognitive Science theory taking pattern recognition as a necessary and sufficient principle. With derived concepts as pattern processing and pattern learning, a complete explanatory model of the mind functioning will be proposed.

This article proposes a set of definitions and explanations as physical phenomenon, of some ill-defined concepts in pattern recognition and cognitive science, actually viewed as abstract principles. The derived explanations and solutions given here is a very ambitious and new scientific proposal.

This paper starts arguing the necessity of the underestimated concept of pattern recognition as a cognitive science principle and the proposal to also be a sufficient principle. Then, a more fundamental description of the pattern recognition concept and it's mechanisms is discussed, allowing a formal physical description and a related cognitive description. This gives a better foundation for the modeling of any cognitive function as a physical pattern recognition mechanism, and explanations to some key biological cognitive processes.

This is a totally new perspective to the pattern recognition concept, and consequently a totally new candidate to be a theoretical solution to some fundamental problems of cognitive science.

The pattern recognition description of some primitive cognitive concepts like instinct, action, reaction, processing and learning are presented. This is important to describe the overall cognitive phenomena in full explanatory level. Also a general unsupervised learning model is proposed, where the only learning conditions are instincts (instinct driven learning).

Then, proposals of more objective and understandable definitions and explanations of controversy cognitive concepts like consciousness are given. This is a proposal for a solution to many historical cognitive science controversies, and the main argument that pattern recognition is also a sufficient cognitive science principle.

As any candidate as a complete theory of cognitive science, this model allow us to re-discuss the philosophy and foundations of science and the human understanding of the universe. With the promising applied pattern recognition technology already in development, this model could help to give some directions to artificial intelligence and also neurobiology and psychology-sociology research.

## 2. Pattern Recognition is a Necessary Cognitive Science Principle

Pattern Recognition is already established as a necessary concept to understand the mind functioning. No scientific study denies that a wide range of cognitive phenomena like detection, perception or sensing are intimately related to pattern recognition processing. Also, intense research in pattern recognition processing in biological, neural, artificial and computational cognitive systems are actually one promising front end of science and technology. But no cognitive science model discuss formally the pattern recognition concept as a general and fundamental cognitive science principle. Such approach is of fundamental importance since pattern recognition processing is one of the mechanisms of mind functioning most observed experimentally. Since actually there is no other technique, formalism or model in the cognitive science that can substitute the pattern recognition functionality, this is a proposed proof that pattern recognition is a necessary and fundamental cognitive science principle.

## 3. Pattern Recognition can be a Sufficient Cognitive Science Principle

It is generally accepted that pattern recognition can explain sensorial and perceptual cognitive functions, but it is not the same for more controversial cognitive functions like consciousness. This paper argues that the same pattern recognition mechanisms can perform instinctive, processing, learning and acting functions. Then it is shown that with a pattern recognition cognitive science formalism we can build definitions an explanations of any controversial cognitive function as consciousness, with an understandable description. With an objective formalism it is possible to propose qualitative and quantitative cognitive science models to be experimentally tested.

All this allows in this paper to claim that pattern recognition can be a sufficient principle to cognitive science.

## 4. Physical and Cognitive Concept of Pattern Recognition

The concept of pattern recognition is widely used in cognitive science related areas. It is generically defined in Math, Computer Science, Design, and few other areas as a method or technique of classification, regularity search, etc. But the formal overall concept of pattern recognition from physical to algorithmic basis is not clear, coherent or unambiguously defined in the scientific literature [Verhagen 1975], [Jie Liu, Jigui Sun, and Shengsheng Wang 2006].

Here is proposed some definitions of pattern recognition first as a general physical phenomenon. This is important because pattern recognition is frequently viewed as an artificial technique rather than a basic natural phenomena. Then is derived the concept of cognitive pattern recognition, and its difference from the basic physical pattern recognition. Also is proposed the not commonly used concepts of pattern processing and pattern learning as playing key cognitive functions to explain the mind functioning.

### 4.1. Pattern and Pattern Recognition – definitions and mechanisms

Here is proposed that the concepts of pattern and pattern recognition can be defined first as a general physical phenomenon:

**<u>Pattern recognition</u> is the physical phenomenon of identification of physical quantities changes of a physical system by a physical pattern recognition mechanism.**

**<u>Pattern</u> is a set of identifiable physical quantities changes of a physical system by a physical pattern recognition mechanism.**

The above definitions depends strictly on the pattern recognition mechanism. An accurate definition of the mechanism is necessary to accurately define the general pattern recognition concept as a physical phenomenon. Also the words identification and identifiable actually belongs to the cognitive science scope, and is generally an abstract concept related to detection, perception, sensing, measuring, etc, functions. The advance proposed here is that the identification cognitive concept can be a well defined physical phenomenon.

The pattern recognition phenomenon occurs when a physical system interacts **changing** some system **physical quantities that can be defined as the pattern recognition identification mechanism**. For example, a lever is a system that can recognize a force applied in one of its side by changing the lever side position, a molecule is a mechanism that can recognize another chemical reactive molecule by changing the molecule chemical structure, a living cell is a system that can recognize many inorganic and organic molecules by metabolic activity, etc.

A conclusion proposed here is that any simple or complex machine, any functional interacting system, or any physical law or interaction, are formally isomorphic to this definition of physical pattern recognition mechanism, as will be argued further below in the philosophical proposal (section 7).

Then some questions arise. How can we relate or differ this basic physical pattern recognition description to the cognitive description of living systems or algorithmic and computational pattern recognition? For example, the immune system can recognize and also memorize a great number of organic molecules and also learn to recognize new antigen patterns and memorize them. Can it be defined as an intermediary pattern recognition system between a basic physical pattern recognition and a neurological cognitive pattern recognition definitions? And what are the key functions of cognitive pattern recognition systems, and do they have any advantage from the more basic physical pattern recognition systems?

A proposed definition of a **Cognitive Pattern Recognition** systems like human mind is the capability of **combinatorial integration** of **processing**, and **learning** of **inter-activating** physical values changes, that can be identified to **control** the overall **behavior**, performing recognition, action-reaction and decision functions. Some advantages of this definition are that combinatorial systems structure scale to an exponential number of possible patterns. A greater number of possible patterns is equivalent to a greater number of possible physical values to be controlled. In other words, can be or build representations of a greater number of systems. Also a greater number of controlling patterns is compatible with a better controlling probability. And mainly, the neural connectionist and artificial intelligence models fits exactly this cognitive pattern recognition definition.

By this criterion the lever is too simple to be considered a cognitive pattern recognition mechanism as it does not have learning and processing like functions. Also, the immune system, even having some learning and memory capabilities, can not be considered a cognitive pattern recognition system because it does not exhibit a mechanism to relate different or previous antibody-antigen patterns in a combinatorial processing scale, according to the standard model of immunological systems.

Many other cognitive science related concepts can have a formal physical and a cognitive definitions. Physical activity can be defined as physical values changes, and activity in real physical systems generally occurs with a dynamic chain or flow of activities, which is the same as to say that physical values changes causes a chain of physical values changes in many physical systems. Physical behavior can be defined as physical activity, and many physical systems of cognitive interest are structured as a combination of small active systems which sum interactions, such as a net force, that is analogous to the concept of control or pattern recognition. This is to argue that variations of equivalent physical and cognitive definitions can be proposed to this area.

A generalization of the physical pattern recognition concept to a more familiar cognitive concept is that **physical quantities can be called properties**. So a simple cognitive pattern recognition definition useful to apply to most related scientific areas, as shown in Table 1 and Figure 1 bellow, is:

**Pattern** is a set of identifiable properties of a system.

**Pattern recognition** is the process of properties identification.

Further in the philosophic discussion (section 7) we propose that every human concept is a set of properties, then a set of patterns.

**Table 1. Definitions of cognitive pattern and pattern recognition concepts as a property concept.**

| SCIENTIFIC AREA | | PATTERN DEFINITION | PATTERN RECOGNITION DEFINITION |
|---|---|---|---|
| **Physics** | | set of identifiable properties of a physical system | event of identification of a pattern by a pattern recognition mechanism |
| **Cognitive Science** | NEUROBIOLOGY | set of properties of active neural cells or synapses | activation of neural cells or synapses |
| | MATH, COGSCI (AI, Neural Nets ) | **a set computable properties** | **functional or computational association of properties** |
| | PSYCHOLOGY | set of behavioral properties | recognition of behavioral properties |
| | PHILOSOPHY | concepts properties definitions | recognition of concepts properties |

In Table 1 and Figure 1, the Neurobiology, AI and Neural-network definitions and mechanisms are suggested according to the dominant scientific interpretation of the connectionist and computational cognitive models. The neural patterns can be modeled either as a pattern of neurons sets or patterns of synapses sets.

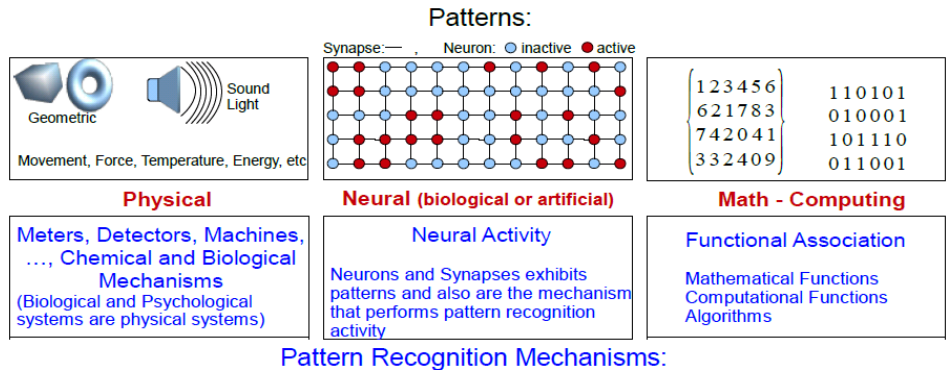

Figure 1. Pattern and pattern recognition concept illustration

## 4.2. Pattern Processing – definitions and mechanisms

In the cognitive science literature, the term pattern processing generally refers to any method, function or algorithm for doing pattern recognition instead of a basic physical phenomenon that can be useful to cognitive science, as is proposed bellow.

When a pattern is recognized by a set of neurons or synapses, for example, a set of retinal cells are activated by a light image pattern, this can be considered as a first step of neural pattern recognition event. The retinal cells can activate a series of synapses though the optical nerve neurons and a set of cortex and other brain areas neurons will be activated in a second pattern recognition event. These inter-activated neural pattern recognition chain, characterizes a processing sequence of pattern recognition events. This suggest us to propose a following definition:

**Pattern Processing is a time step or sequential chain of inter-activated pattern recognition events.**

The same definition can be applied to any, physical, mathematical, computational, artificial intelligence or neural network pattern recognition systems or method. In principle, any algorithmic implementation of pattern recognition can be used sequentially or recurrently to others to perform pattern processing. This occurs in the computational pattern recognition implementations and possibly occurs in the biological neural systems.

This notion of pattern processing is not usual in the cognitive science literature. Most of pattern recognition scientific studies are related to perception or sensory concepts or mechanisms. The considered "higher order cognitive processing mechanisms" like, thinking, consciousness, etc, are frequently discussed as a different processing mechanism from the original pattern recognition functional mechanism. In the connectionist models, there is also a discussion about integrative functions, global workspace functions, synchronization, etc, but is not clear in those models an explicit reference to an overall and strict neural pattern processing function.

The important new concept proposed here is that **the same functional mechanism that performs pattern recognition can perform pattern processing**. This notion can be very useful because it can be applied to biological, artificial and theoretical pattern recognition and processing studies, and can be also a unified cognitive concept to define and build explanations to supposedly any cognitive phenomena, as is proposed in this paper.

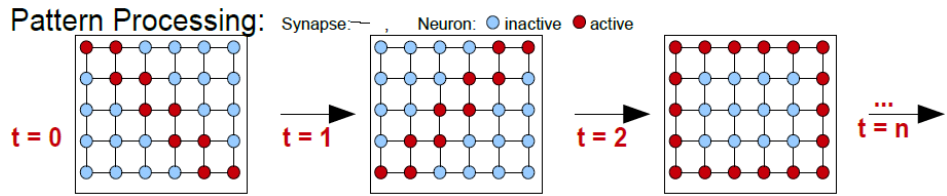

Figure 2. Pattern processing concept illustration

As the same as the neural network can be considered to performs the same role of standard computation [McCulloch and Pitts 1943], the pattern processing can equivalently perform the two basic computational instructions:

DO A – when a pattern is activated in the pattern recognition mechanism by default, without a previous sensory or processing activation pattern. Biological analogy could be any self activated neural pattern as apparently is the sleepy instinct. An artificial analogy can be any preprogrammed instructions or algorithmic function not or less dependent from others instructions causing patterns to be activated independently of any other pattern as are for example the computer clock dependent instructions.

IF A THEN B – the basic pattern recognition concept when a pattern A (sensorial, processing or self activated) is active and then it activates another pattern B. This is the usual notion of pattern recognition in the detection function where a external physical pattern activates a related sensorial pattern A that activates, or is recognized, to a correspondent pattern B in the pattern recognition processing mechanism.

### 4.3. Pattern Learning – definitions and mechanisms

As the same as the concept of neural network synaptic weight functions changes can be interpreted as biological neural learning, we can have a general definition of pattern learning as:

**Pattern learning is the changing of pattern recognition identification set.**

This definition is useful because it is equivalent to the intensive studied standard neural connectionist and artificial intelligence learning models. Also because it can be a generalization of these learning mechanisms as a basic pattern recognition process as learning been simply the recognition of new patterns.

A simple example can be the recognition of a new pattern of numbers sequence when a person dial a phone, or recognition of a new face image pattern to a new person.

A consequence of this definition is that the concepts of learning and memory are equivalent. When we memorize we are learning new patterns, and when we learn we are memorizing new patterns. This equivalence is not always clear in the cognitive science literature.

A difference between learning and memorization can be established if one define different categories of patterns as for example simple recorded patterns of known processing patterns as memory and new processing patterns recognition not previously known as learning, but it is only a matter of categorization of the basic mechanism of pattern recognition.

A conclusion we can propose is that **the learning concept is a subset of the pattern recognition concept**.

## 5. Pattern Recognition definitions and explanation of cognitive phenomena

The above physical to cognitive definitions and discussions are a formal basis to propose a general theory for cognitive science in the following steps:

- Describing instincts as pattern recognition functional mechanism.
- Proposing an unsupervised instinctive conditioned pattern leaning model.
- Proposing pattern recognition definitions and explanation mechanisms of any higher order cognitive concepts as thinking and consciousness.
- Applications proposals.

### 5.1. Instincts, Sensing, Feelings and Emotions - Basic Cognitive Phenomena

The basic animals instincts concept can be proposed to be functionally isomorphic to the pattern recognition concept, and pattern recognition mechanisms as proposed here can be a mechanistic description of any instinctive related cognitive function.

As far as the patterns, or set of properties, of one or a set of instincts can be described, a pattern recognition processing can be proposed as a functional mechanism. A simplified proposed description follows.

Basic instincts as pain, fear, desire, surprise, etc, can be described as predominantly built in pattern recognition functions that are relatively independent from learning and can have a great influence or even dominate the overall pattern processing at a period of time. The expression predominantly built in was used assuming that most biological instincts are considered mainly biological basic functions with less influence from phenotype learning, even some instincts seems able to learn to react somehow differently as for pain tolerance training, learned traumas, etc.

A simple explanation to the biological instinct mechanisms can be analogous to an artificial pattern recognition processing by establishing some previous categories of instinct activation patterns like:

Action Instincts – "DO A" activation patterns – when a pattern or a set of patterns are preprogrammed to be active at periods of time, performing an analogous role of computer "processing clock" functions. The neural analogous can be sets of neurons which fires synapses periodically without any previous stimulation, playing the role of "neural system clock". These independent neural activations can be due to any neural or body metabolism or structure, being less dependent on the synaptic mechanisms as a source, but then doing standard synaptic role as an input pattern in the pattern recognition mechanism. Possible biological "DO A" instincts are sleeping, awaking, hungry, compulsive or involuntary muscle motion, and even a explanatory basis for attention, will, desire and others cognitive functions.

Reaction Instincts – "IF A THEN B" activation patterns – when a reaction to a category of activated input or initial patterns have a predefined response or processing pattern. The computational analogous can be any set of predefined instructions for a input pattern category. Possible neural analogous can be any predefined reaction from stimulus like pain reactions, scare reactions, reflex, etc.

Sensing, detection or simple recognition is a cognitive function strict related to the pattern recognition concept. Many pattern recognition methods are inspired in biological recognition similarities. Feelings and emotions can be simply described as instinct reactions patterns associated to learned (cultural) patterns.

The role of instinct patterns in this theoretical proposal is to be the initial, contour and ongoing conditions for pattern processing and learning.

## 6. Processing and Learning Strategies based on Instinctive Patterns

Another proposal is that the overall cognitive processing activity can be consistent with the Pattern Processing concept as defined and discussed in section 4.2 as a chain of inter associated cognitive pattern recognition processing activations or occurrences. The advance that this proposal can give to Cognitive Science is to identify some cognitive meaning from the great amount of processing activities of the cognitive systems, either biological neural systems or artificial intelligence. This is also a proposal for an overall solution for the cognitive representation problem, since it objectively defines the cognitive representation structure where the mechanisms are pattern recognition and processing mechanisms and the format structures are patterns.

Based on the definitions and discussions of the section 4.3 we can propose that the cognitive learning mechanisms can be also described as a pattern learning mechanisms. A simple unsupervised learning strategy proposed here to describe the overall cognitive learning capability is the criteria that instinctive patterns are the conditions for pattern learning.

The following examples are simple descriptions of:

1. Baby Motor Learning – the muscle movement learning of babies can be described simplified as driven for the instinctive patterns of innate muscle random movement and the sensations of pain, discomfort, relief, equilibrium etc. Since born the babies instinctively moves their muscles randomly with increasing amplitudes and so they learn positions that avoids pain and discomfort giving optimized equilibrium, stability and relief for stress conditions like itch.

2. Baby Optical Learning – in the same way, instinctive random visual search and stability and relief conditions can be a simple description for optimized visual learning.

3. Visual and Motor Learning – the time correlation events of motor and visual learning gives the visual-motor common optimized recognition. This can be a simple description of the concept "global cognitive workspace integration or coherence" frequently cited in cognitive science discussions.

4. All human learning capabilities can be explained similarly.

The specific instinct learning model may be various from the AI and neurobiology literature. One simple and possibility universal learning strategy consistent to this proposal is the positive and negative reinforcement based on stress relief, which can be defined as awareness. Known process consistent to the notion of instinctive driven pattern learning is Pavlovian behavioral learning and associative learning.

## 6.1. Self, thinking and consciousness pattern recognition definitions

This section gives a set of proposed definitions and explanations as an argument that Pattern Recognition is a sufficient principle to Cognitive Science. A very convincing argument is that it is a natural formal representation language for defining controversial cognitive concepts like self, thinking and consciousness, and also the only known scientific established physical process concept able to give universal understanding of any cognitive phenomenon.

And more, any set of patterns required to define or to explain any cognitive phenomenon can in principle be processed by any suitable physical pattern recognition mechanism. If such a set of patterns shows to be insufficient to describe any aspect of the phenomenon, either behavioral, neurological or philosophical, then those aspects can be added to the definition pattern set. Bellow is proposed some Pattern Recognition cognitive definitions, intended to be a first universal proposal, but that can be further discussed and fitted according to the vast knowledge of the many branches of cognitive science related fields.

Some simple cognitive concepts definitions and explanations based the pattern recognition principle follows. The cognitive definition or description of the concept of self we can propose is simply the overall set of patterns related to the geometrical limits of one cognitive system activities. For a human, we can recognize the sets of patterns that are related to our body structure, our body activities and interactions, in contrast to others patterns that do not interacts directly with us.

The human **thinking activity** can be defined or described as **the commanding pattern recognition, processing and learning cognitive system**. This definition assumes that there exists cognitive activities that are relatively independent from the commanding thinking processing like surprise or pain reactions that generally occur without any thinking premeditation. Generally the thinking systems only recognizes this kind of activities after it occurs. Examples of cognitive activities that apparently are not recognized and processed by the cognitive thinking are some facial expressions that we exhibit during communication, the same with the complex muscle movement patterns of a simple walk and we are not always aware, or we simply do not recognize, all movement steps and decisions, and many others cognitive activities we are not even aware we perform them. On the other side many actions and decisions are controlled and recognized by the thinking processing and theses generally commands the overall actions, they forms a chain of cognitive commanding process that we can verbally report, plan, etc.

In the cognitive thinking processing we can generally premeditate or preprocess many activities or actions patterns, these plans are also recognized by the thinking processing. Then the cognitive thinking processing can execute the planned activities, also this execution are recognized by the thinking processing, and after the set of activities are already done, they are recognized to have occurred. In all these steps the cognitive thinking processing commands the overall activities and can for example report it verbally or with any other communication tool. In some cases the thinking process only recognize and process a pattern after it occurred, without premeditation as it occurs in surprise.

With this we can propose a definition of **consciousness as the cognitive pattern recognition activities recognized by the thinking system**. Simply we can describe any conscious activity as any activity recognized by a pattern recognition commanding process. Self consciousness as the thinking recognition of the patterns of its own activities. Self cognitive consciousness as the recognition of the patterns of its own cognitive activity. A statement to summarize this proposal is that a sufficient powerful pattern recognition system may be able to recognize the patterns of it's own activity, including the cognitive activity.

These are objective, understandable and unambiguous construction of a cognitive science definition to the word consciousness. For example, by this definition a computer where we can define a commanding system as its operational system, OS, is conscious of any pattern that the OS recognizes, as a file, a process, etc. Also as the SO can recognize many patterns of its own activity, as it's memory or disk usage, cpu temperature, peripherals device patterns, etc, the OS is self conscious of many of its own patterns. In the same way an animal have self consciousness of its pain, hungry, etc , patterns.

## 7. From applications to philosophy

As the pattern recognition scientific area is very broad, promising and active, we can expect that many above proposals can have applications, and the fast development of pattern recognition science and technology is a serious candidate to answer many cognitive science questions.

One possible application is that objective quantitative criteria for non human conscious and unconscious recognition can be proposed. A two level commanding or controlling criterion for a specific subset of pattern recognition to be considered conscious could be statistical or probabilistic. For example, any pattern that have a probability greater than 98% to be recognized for an artificial intelligence or animal study could be defined as conscious, and patterns with probability less than 2% can be considered unconscious. Or even can be proposed a graduation of conscious pattern recognition probability. Probability distributions can also be a proposed criteria, as for any specific function or geometry as for sharp Gaussian distributions, the conscious can be the peak interval and the tails the unconscious.

Other possible criteria for non human conscious recognition could be communication, where only patterns that can be explicitly be reported by a specifically defined communication process could be accepted as conscious, or by learning criterion, where the patterns before a learning process are unconscious and conscious after learned. But it is evident that the whole discussion at this paper points that any application of the consciousness concept is strictly related to the pattern recognition concept.

Other application of the pattern recognition concept in this paper is that any human concept can be defined as a set of properties. So by the definitions in this paper, any human concept is a set of pattern, and a pattern recognition mechanistic process can be proposed as a general or the basic explanatory formalism to philosophy. About the philosophy and foundation of science, the general and basic scientific representation of any scientific field can be proposed as a pattern recognition formalism.

A question that arises is, are there basic physical or cognitive patterns that spans any other compost pattern? Physics basic quantities are the most fundamental patterns in natural science. By the cognitive pattern recognition concept proposed here, the most basic pattern is physical quantity change, the same as interaction or activity. From the connectionist point of view, the most basic and natural pattern representation is the binary representation, and any other pattern representation is isomorphic to a binary representation.

The cognitive structure and mechanisms can be derived by the basic physical quantities. But reversely, the we can propose that the physical quantities can be derived by the pattern recognition concept. A pattern recognition structure, like neural connectionist, can be proposed to recognize a geometrical point, then many geometrical points, then to have distinct recognition reactions to different point numbers, or say counting and then metrics, recognize linear, curve, plane and solid dense point distributions, or say geometry.

So another interesting proposal we can discuss is that a physical cognitive explanatory model can suggest hints to discuss the cognitive basics of natural science, since many fundamental concepts of science are viewed as basic or primitive cognitive or intellectual properties, like the dot geometrical concept can be cognitively viewed as a pattern recognition property.

## 8. Conclusions

This paper proposes a new solution to the problem of the ill-defined pattern recognition concept [Verhagen 1975], [Jie Liu, Jigui Sun, and Shengsheng Wang 2006], by defining pattern recognition as a physical phenomenon. This brings the principles of the pattern recognition scientific field with a concrete physical basis other than an abstract ambiguous ground.

This paper also proposes a solution to the cognitive science core problem of establishing unambiguous and understandable definitions, explanations and descriptions of controversial cognitive science concepts as consciousness. Taking the principle of pattern recognition as a formal explanatory concept we showed the possibility to provide physical and concrete explanations of the processing mechanism of the mind functioning, giving an objective solution to many difficult and abstract questions on the field.

All the set of conceptual mechanistic explanations and definitions proposed in this paper can be compared to others similar key scientific proposals. The Darwin Evolutionary Theory was proposed also as a set of conceptual explanations and definitions proposals. At the time of Darwin the biology had even less established knowledge than pattern recognition field has today. Also the original Hebbian proposals of neurological learning mechanisms had less theoretical, experimental and applied basis than pattern recognition actually. The importance of these two previous conceptual scientific proposals is an example in support to the validity of the conceptual proposals made in this paper.

All discussions in this article gives the conviction to claim that pattern recognition is a sufficient explanatory principle to cognitive science, and thinking, consciousness and recognition are strictly related concepts.

This paper is submitted to peer review publication (May 2011).

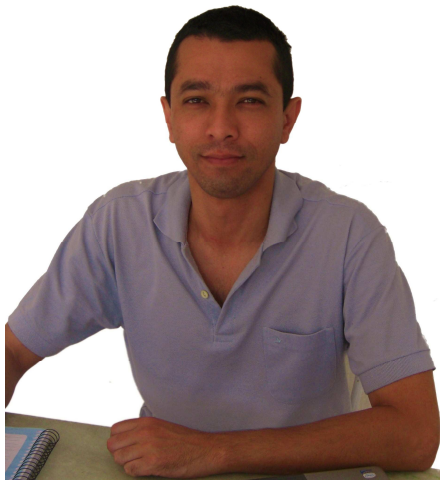

Gilberto de Paiva